\documentclass{article}

\usepackage{PRIMEarxiv}

\usepackage[utf8]{inputenc} 
\usepackage[T1]{fontenc}    
\usepackage{hyperref}       
\usepackage{url}            
\usepackage{booktabs}       
\usepackage{amsfonts}       
\usepackage{nicefrac}       
\usepackage{microtype}      
\usepackage{lipsum}
\usepackage{fancyhdr}       
\usepackage{graphicx}       
\graphicspath{{media/}}     
\usepackage{graphicx}
\usepackage{booktabs}

\usepackage{amssymb}
\usepackage{graphicx}
\usepackage{array}
\newcolumntype{C}[1]{>{\centering\arraybackslash}p{#1}}
\newcolumntype{L}[1]{>{\arraybackslash}p{#1}}
\newcolumntype{R}[1]{>{\raggedleft\arraybackslash}p{#1}}

\usepackage{algorithm}
\usepackage{algorithmic}
\usepackage{amsmath}

\usepackage{tabularx} 
\usepackage{booktabs} 

\usepackage[accsupp]{axessibility}
\title{Point Mamba: A Novel Point Cloud Backbone Based on State Space Model with Octree-Based Ordering Strategy
}

\author{
  Jiuming~Liu$^{\dagger}$ \\
  Shanghai Jiao Tong University  \\
   \And
  Ruiji~Yu$^{\dagger}$ \\
  Shanghai Jiao Tong University \\
  \And
  Yian~Wang \\
  Shanghai Jiao Tong University \\
  \And
  Yu~Zheng \\
  Shanghai Jiao Tong University \\
  \And
  Tianchen~Deng \\
  Shanghai Jiao Tong University \\
  \And
  Weicai~Ye \\
  Zhejiang University \\
  \And
  Hesheng~Wang \thanks{Corresponding Author. The first two authors contribute equally.}\\
  Shanghai Jiao Tong University \\
}

\begin{document}
\maketitle

\begin{abstract}
Recently, state space model (SSM) has gained great attention due to its promising performance, linear complexity, and long sequence modeling ability in both language and image domains. However, it is non-trivial to extend SSM to the point cloud field, because of the causality requirement of SSM and the disorder and irregularity nature of point clouds. In this paper, we propose a novel SSM-based point cloud processing backbone, named Point Mamba, with a causality-aware ordering mechanism. To construct the causal dependency relationship, we design an octree-based ordering strategy on raw irregular points, globally sorting points in a z-order sequence and also retaining their spatial proximity. Our method achieves state-of-the-art performance compared with transformer-based counterparts, with 93.4\% accuracy and 75.7 mIOU respectively on the ModelNet40 classification dataset and ScanNet semantic segmentation dataset. Furthermore, our Point Mamba has linear complexity, which is more efficient than transformer-based methods. Our method demonstrates the great potential that SSM can serve as a generic backbone in point cloud understanding. Codes are released at https://github.com/IRMVLab/Point-Mamba.
\end{abstract}


\section{Introduction}
\label{sec:intro}
Point cloud is a fundamental 3D representation, which provides continuous spatial position information with XYZ coordinates. It has a wide range of applications in many fields, such as autonomous driving \cite{liu2023translo}, virtual reality \cite{wang2023pointshopar}, and robotic vision \cite{seita2023toolflownet}, etc. 

However, the intrinsic disorder and irregularity natures of point cloud challenge the effective point cloud processing methods. With the booming development of deep learning, recent works basically focus on the learning-based point cloud understanding methods. They can be mainly classified into two categories: \textit{voxel-based methods} and \textit{point-based methods}. Early voxel-based methods use uniform voxel sampling to extend the convolution operation of 2D CNN to 3D space \cite{2,3}. However, the voxel-based methods suffer from low resolution and high computational overhead due to their cubic complexity \cite{1,2,3,4}. To overcome the bottlenecks, sparse-voxel-based methods apply convolution operations in the sparse non-empty voxels and introduce some tricks such as octrees and hash tables to more effectively organize the point cloud \cite{15,16,17,18}. However, the sparse convolution operation utilizes the small convolution kernel on a limited voxel size, which fails to capture global features\cite{15}.

Point-based methods, which are pioneered by PointNet and PointNet++ \cite{10,11}, directly learn point features from raw point cloud data. Typically, point-based methods have two common operations: First, to aggregate neighborhood features, they construct the k-nearest neighbor searching (KNN). Second, to extract hierarchical features, farthest point sampling (FPS) \cite{10} or gird sampling \cite{71} is introduced to downsample the point cloud. However, both KNN and FPS have an expensive computational overhead, which limits the point-based methods for generalization on large-scale point cloud datasets. Furthermore, early point-based methods are typically established on CNN with a limited receptive field. To enlarge the receptive field, the architecture of the transformer \cite{54,48,55} backbone has been recently introduced into point cloud processing. Among them, PCT \cite{22} utilizes the self-attention mechanism to capture the global features. Point Transformer \cite{24,25,26} uses a progressive downsampling followed by vector attention to extract aggregated local features. Considering the quadratic complexity of the transformer \cite{43,54}, recent transformer-based works reduce the computational overhead by partitioning window \cite{58} or applying the voxel-hash method \cite{28} to sparse the attention matrix. However, these attempts will sacrifice the global receptive field to some extent.

Recently, a new backbone Mamba, which is based on the state space model (SSM)\cite{43} has achieved both linear complexity and the long-range context learning abilities in sequence data. Mamba innovatively introduces hardware acceleration and selective scanning mechanisms and has the potential to serve as an alternative to the transformer backbone in language modeling and image processing \cite{43,49,50}. Nevertheless, a general SSM-based backbone has not been investigated in the field of point cloud processing. Inspired by the linear complexity and global modeling capabilities of Mamba, we propose a novel backbone named Point Mamba for 3D point cloud understanding.

However, introducing Mamba into point cloud processing still has a great challenge: The original design of the Mamba backbone is aimed at solving casual sequential language tasks \cite{43}, which obey the inherent non-casual nature of point cloud data. To solve this problem, we design an octree-based ordering mechanism inspired by OctFormer \cite{45} to construct the causal dependency relationship based on a z-order curve for sorting the irregular point cloud data. Octree nodes are sorted by the shuffled keys, which can retain the original spatial proximity of raw point clouds. Then, we embed point features in different downsampling stages to get the hierarchical features for downstream tasks of the point cloud.

In summary, the main contributions of our work are as follows:
\begin{itemize}
  \item We propose a novel point cloud backbone, named Point Mamba, based on the state space model (SSM) for various point cloud understanding tasks.
  \item To solve the non-casual issue, we introduce an octree-based ordering scheme to reconstruct the point cloud order and form a causal dependency relationship based on the z-order curve.
  \item Our method achieves 93.4\% classification accuracy on ModelNet40 and 75.7 mIOU on the large-scale semantic segmentation dataset ScanNet, which exceeds the transformer-based counterparts. Also, Point Mamba has linear complexity, enabling high point processing efficiency.
\end{itemize}
  


\section{Related Work}

\subsection{Generic Point Cloud Processing Backbone:}
Previous learning-based point cloud processing methods are mainly established on two fundamental backbones: CNN and transformer, which have substantially advanced the research in point cloud processing.

\noindent\textbf{CNN-based Backbone.}\hspace*{5pt} Early works on point cloud processing mainly focus on applying CNN to 3D point cloud data. Some works \cite{59,60} project the point clouds to 2D images and then apply the 2D CNNs. However, this projection process will fold the spatial information of the point cloud, leading to the inevitable information loss \cite{24,25,26}. Other works directly apply convolution operations on point clouds in 3D space, which can be basically divided into two categories: voxel-based methods \cite{1,2,3,4} and point-based methods \cite{5,6,7,8}. Although 3D CNN can retain more spatial information about the point cloud compared to the projection method, the limited local receptive field of the CNN backbone makes it difficult to capture global features.

\noindent\textbf{Transformer-based Backbone.}\hspace*{5pt} Compared with CNN, transformer has an advanced advantage of capturing global features, due to the global attention mechanism \cite{54,48}. PCT \cite{22} 
enhances input embedding with the support of farthest point sampling and nearest neighbor search and uses self-attention to capture global features.
PointASNL \cite{23} applies an adaptive sampling module and a local-nonlocal module to balance global and local features. Point Transformer \cite{24, 25, 26} uses progressive downsampling followed by vector attention to extract aggregated local features, and Fast Point Transformer \cite{28} applies voxel hashing-based architecture to enhance computational efficiency. However, the quadratic time complexity of attention calculation constrains their efficiency, especially for large-scale scalability \cite{48}. Recently, some works \cite{31,32} have explored improving point transformers with window attention inspired by Swin Transformer \cite{73}. For example, SST \cite{31} and SWFormer \cite{32} calculate the attention only within local partitioned windows. However, the window partition process degrades the global modeling ability of transformer. Recent success of Mamba based on the state space model (SSM) provides a new perspective to solve the above bottleneck of point transformer backbone \cite{43,49,50}, due to its long-sequence modeling and linear complexity abilities \cite{43}.

\subsection{State Space Model (SSM) and Mamba}
\noindent\textbf{State Space Model.}\hspace*{5pt} State space model (SSM) is initially a mathematical model used to describe dynamic systems in modern control theory. It is recently introduced as a generic backbone in natural language processing and computer vision domains \cite{37,38,41}. By introducing HiPPO \cite{36} initialization, LSSL \cite{37} showcases the potential to handle long-range dependencies for sequential data. However, LSSL is not practical due to the computational and memory overhead. S4 \cite{38} introduces the normalization of parameters into diagonal structure to solve the problem of LSSL. More attempts are proposed to further improve the structure of SSM, such as complex-diagonal structure \cite{39,40}, multiple-input multiple-output supporting \cite{41}, and decomposition of diagonal plus low-rank operations \cite{42}. These works are applied to large representation model \cite{44,45} and mainly focus on how state space models are applied to casual data like language and speech \cite{43}. Some works also extend the SSM model to image data, such as pixel-level 1-D image classification \cite{38}. S4ND \cite{47} is the first work applying state space mechanism to visual tasks and shows the potential to compete with ViTs \cite{48}. However, the previous language processing SSM models fail to efficiently capture information in an input-variable manner, because they all expand the S4 model in a simple way \cite{43}. The problem of these early SSM models is similar to RNN models, which means that these models will forget global-context information and have gradient vanishing problems as the sequence length grows.

\noindent\textbf{Mamba.}\hspace*{5pt} With the characteristics of linear complexity and selective scanning mechanism, Mamba \cite{43} has the potential to be a novel backbone that can replace the transformer backbone in language modeling. Furthermore, Vision Mamba \cite{49} and Vmamba \cite{50} successfully apply Mamba to the field of image processing through bidirectional selective scanning and cross-selective scanning mechanisms respectively. Our work further expands the application of Mamba into the point cloud field. Through an octree-based ordering scheme, we reconstruct the point cloud order and form the causal dependency relationship.

\section{Method}
In this section, we begin with the description of the preliminaries of the state space model in Section \ref{sec:preliminaries}. Then, we illustrate the overall architecture designs of our method in Section \ref{sec:arch}. For the causality establishment, we introduce an octree-based ordering scheme in Section \ref{octree}. Finally, we will delve into the designs of Point Mamba block in Section \ref{sec:point mamba} and analyze the efficiency of our methods in Section \ref{sec:eff}.

\subsection{Preliminaries}
\label{sec:preliminaries}
\noindent\textbf{State Space Model.}\hspace*{5pt} State Space Model (SSM) is considered as a linear time-invariant system that maps the input sequence $x(t) \in \mathbb{R}^L$ to the output sequence $y(t) \in \mathbb{R}^L$.
Mathematically, it is described by a set of ordinary differential equations (ODEs) as:
\begin{equation}
  \begin{aligned}
    \dot{h}(t) &= Ah(t) + Bx(t), \\
    y(t) &= Ch(t) + Dx(t),
  \end{aligned}
  \label{eq:1}
\end{equation}
where $h(t) \in \mathbb{R}^N$ is the hidden state.
$\dot{h}(t) \in \mathbb{R}^N$ is the derivative of the hidden state.
$A \in \mathbb{R}^{N \times N}$, $B \in \mathbb{R}^{N \times L}$, $C \in \mathbb{R}^{L \times N}$, and $D \in \mathbb{R}^{L \times L}$ are the parameters of the model.

\noindent\textbf{Discretization and Global Convolutional Kernel.}\hspace*{5pt} The continuous-time SSM can be discretized to a discrete-time SSM by zero-order hold (ZOH) discretization. The parameters $A$ and $B$ of the discrete-time SSM can be obtained by introducing the sampling step $\Delta$ and Taylor expansion. In this case, parameters can be approximated as:

\begin{equation}
  \begin{aligned}
    \bar{A} &= e^{\Delta A}, \\
    \bar{B} &= (I - e^{\Delta A})A^{-1}B, \\
    \bar{C} &= C, 
  \end{aligned}
    \label{eq:3}
\end{equation}
where $\bar{A}$, $\bar{B}$, and $\bar{C}$ are the parameters of the discrete-time SSM. In this case, the formula \ref{eq:1} is rewritten as:
\begin{equation}
  \begin{aligned}
    h_{k} &= \bar{A}h_{k-1} + \bar{B}x_{k}, \\
    y_{k} &= \bar{C}h_{k}.
  \end{aligned}
  \label{eq:4}
\end{equation} 
Finally, the model output $y$ can be represented as the convolution of input $x$ and a convolution kernel, which makes Mamba have the potential of parallel computation as:
\begin{equation}
  \begin{aligned}
    y = \bar{K} \ast x,
  \end{aligned}
  \label{eq:6}
\end{equation}
where $\bar{K}$ is the convolution kernel, which can be represented by:
\begin{equation}
  \begin{aligned}
    \bar{K} =(C\bar{B},C\bar{A}\bar{B},... ,C\bar{A}^{M-1}\bar{B}).\\
  \end{aligned}
  \label{eq:5}
\end{equation}

\noindent\textbf{Selective Scanning Mechanism.}\hspace*{5pt} Diverging from the previous approaches that predominantly focus on linear time-invariant (LTI) SSMs, Mamba introduces a selective scanning mechanism to compress the context-awareness of the data. Selective scanning mechanism means that the matrices $B \in \mathbb{R}^{B \times L \times N}$, $C \in \mathbb{R}^{B \times N \times L}$, $\Delta \in \mathbb{R}^{B \times L \times D}$ are derived from the input sequence $x \in \mathbb{R}^{B \times L\times D}$.
This makes Mamba aware of long-distance dependencies in the input sequence contextual information and ensures the dynamic range of weights within the model.
\subsection{Overall Architecture}
\label{sec:arch}
\begin{figure}[tb]
  \centering
  \includegraphics[width=1.0\textwidth]{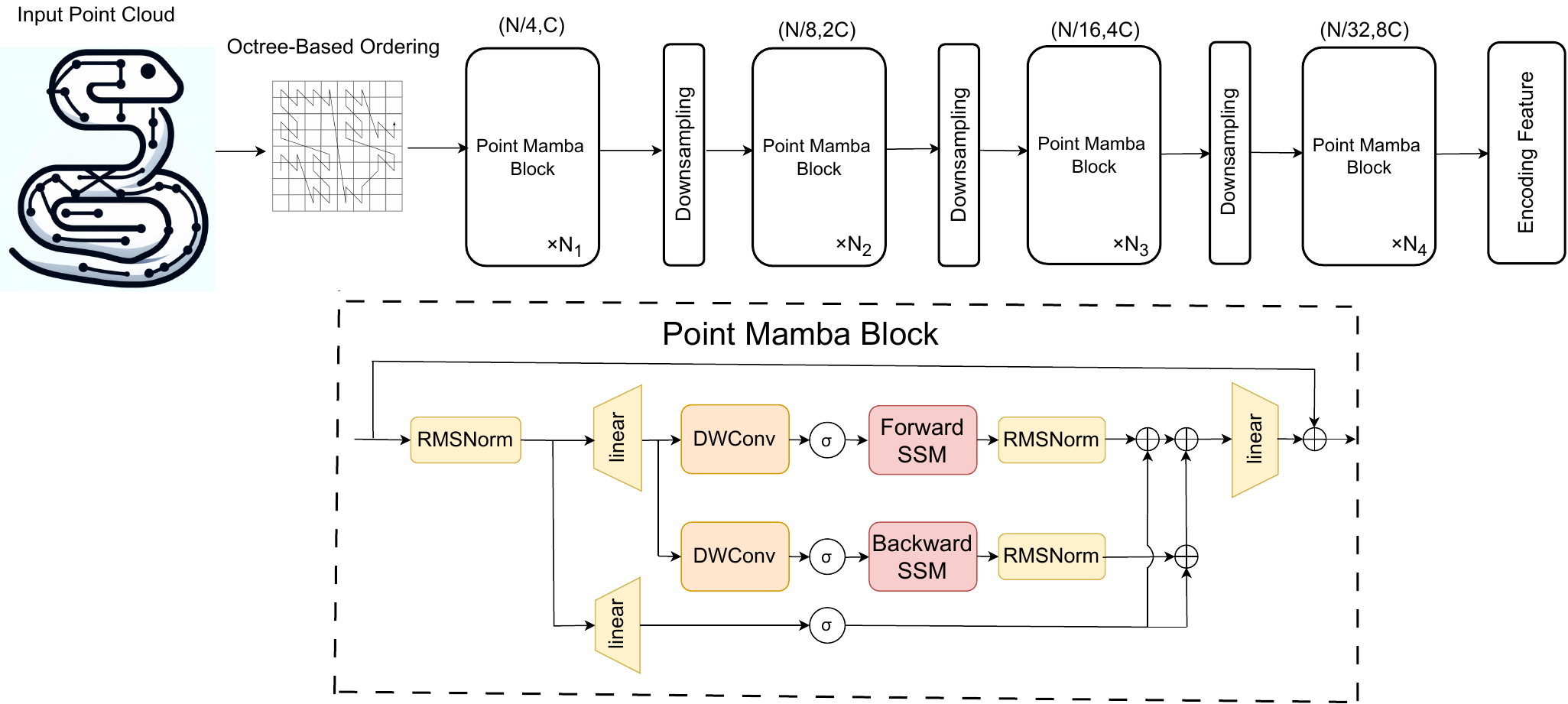}
  \caption{The overall architecture of Point Mamba, which contains an octree establishment layer, a feature embedding layer, and a sequence of Point Mamba blocks and downsampling layers. N is the number of input points. C is the channel dimension of the point features. $N_{i}$ denotes the number of Point Mamba Blocks in the $i$-th stage. The structure of one Point Mamba Block includes the core SSM module and the bidirectional selective scanning mechanism \cite{49}.
  }
  \label{fig:0}
\end{figure}

The overall architecture of our Point Mamba is shown in Fig. \ref{fig:0}. Point Mamba captures the global point-wise features with a hierarchical architecture, which can characterize the semantic affinities between points for various point cloud processing tasks. 

Specifically, We first normalize the input point cloud and build the octree based on it. The point-wise features all are stored in the octree nodes, which are sorted by shuffled keys to form a z-order-based sequence as in Section \ref{octree}. Then, we embed the point cloud features into a higher dimension and apply a series of Point Mamba Blocks in Section \ref{sec:point mamba} and downsampling layers to obtain the hierarchical point features. After getting the output features $P_{out}$, we apply a lightweight Feature Pyramid Network (FPN) \cite{69} for downstream classification or segmentation tasks.

\subsection{Octree-based Ordering Mechanism}
\label{octree}
The most significant challenge of applying the SSM backbone to point cloud processing is how to establish the causality on original disordered point cloud data. Recently, some works \cite{66,27} have discovered that permutation invariance and nearest neighbor searching are not necessary for point cloud analysis. With certain serialization patterns on the raw point cloud, their models can also achieve promising results on a series of downstream tasks \cite{66,27}. Motivated by their success, we introduce the octree-based ordering mechanism in our Mamba block to obtain the causality. \textit{Notably, our ordering strategy is directly utilized to the raw input points, rather than the local patches aggregated from KNN.}

\begin{figure}[tb]
  \centering
  \resizebox{1.0\textwidth}{!}{
   \includegraphics[width=1\linewidth]{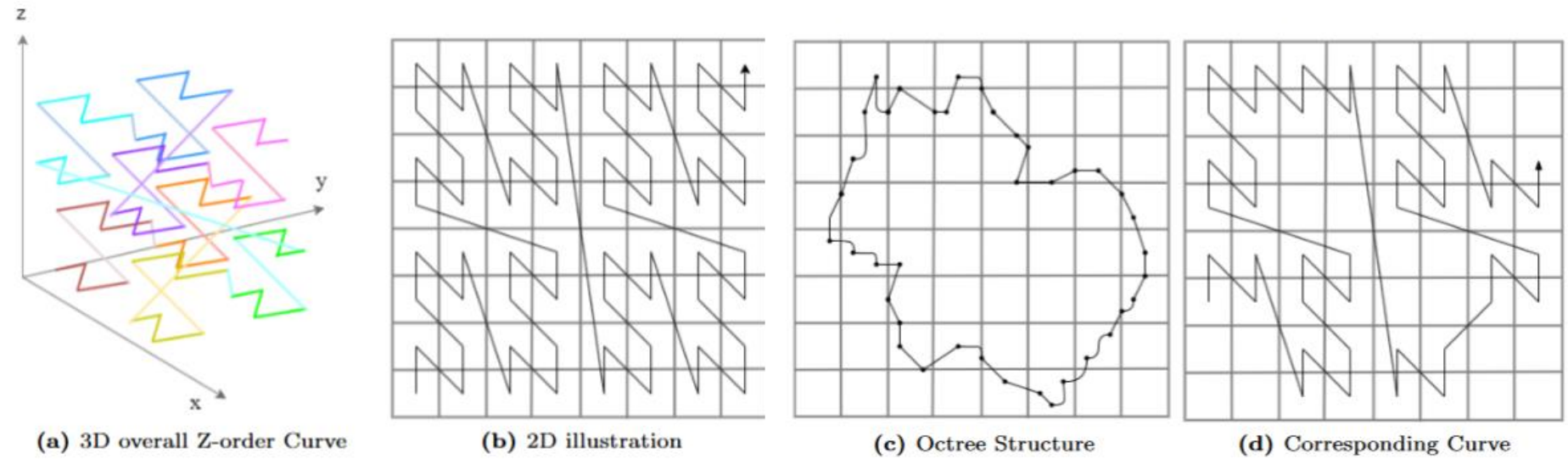}
   }
  \caption{The 3D overall z-order curve and a 2D illustration of octree structure and its corresponding curve. (a) is the overall z-order curve in 3D space, (b) is the 2D illustration. (c) is the octree structure, and (d) is the corresponding z-order curve of the octree structure in the depth of 3.}
  \label{fig:1}
\end{figure}

Specifically, for each point cloud, we first construct an octree structure on GPU \cite{67}. The octree nodes in the same depth are sorted by shuffled keys \cite{68} in binary expression as follows:
\begin{equation}
  \begin{aligned}
    \text{Key(x,y,z)} =x_0y_0z_0x_1y_1z_1...x_{d-1}y_{d-1}z_{d-1},
  \end{aligned}
  \label{eq:7}
\end{equation}
where $x_i$, $y_i$, and $z_i$ represent the $i$-th bit of the coordinate of the point in the octree node and $d$ is the depth of the octree. The value of the key corresponds to the position of the 3D z-order curve, whose 2D illustration is shown in Fig. \ref{fig:1}.

This organized point cloud structure possesses two advantages: one is that the causality relationship derived from the 3D z-order curve is better adapted to the Mamba backbone. The other is that the z-order position preserves the spatially adjacent relationship of the point cloud and is stored accordingly in memory. These two key points make our Point Mamba effectively capture both causality dependencies and geometric information from raw point cloud data.

\subsection{Point Mamba Block}
\label{sec:point mamba}
After the octree-based ordering, input point cloud data is ordered causally and transformed into a 1-D sequence, which is then fed into the Point Mamba block. In this section, we introduce the specific designs of our Point Mamba Block, which combines the bi-directional selective scanning mechanism \cite{49} to adjust the sequence-order dependence of Mamba. 
We first stack the reordered points into a 1-D sequence $P_{in} \in \mathbb{R}^{ N\times C}$, where C is the embedded feature dimension 
and N denotes the number of points.
Considering the linear complexity of Mamba, we can directly process the point cloud sequence with our designed Point Mamba Block without partitioning the window as previous transformers \cite{31,32}. 

Specifically, the algorithm of our designed Point Mamba Block is given in Algorithm \ref{alg:1}, where E is the embedded feature dimension, and M is a fixed parameter, set to 16 by default.

In the Point Mamba Block,
$\Delta_o$ is the control parameter for the discretization, which determines how much the model learns from the current input point\cite{43}. Specifically, if $\Delta_o$ is large, the discretization of the SSM model is low, and the model will learn more about the current input point cloud features. Conversely, if $\Delta_o$ is small, the model will learn more about the global input point cloud features.
Once $\Delta_o$ has selectivity, $A_o$, $B_o$, and $C_o$ will also learn the features of the input point cloud selectively.
As directly learnable parameters, $B_o$ controls the level to which the model is affected by the current input point cloud, and $C_o$ controls the level to which the model is affected by the historical input point cloud.

The core of the Point Mamba Block is a bi-directional selective scanning mechanism, which further forms the causality adapted to Mamba by traversing the two possible orders of the input 1-D sequence. According to the process of Algorithm \ref{alg:1}, we can get $y_{forward}$ and $y_{backward}$ respectively. The two sequences are then gated by $z$ and summed. Finally, with a residual connection and a linear layer, we get the output sequence $P_{out}$.

\begin{algorithm}[tb]
  \caption{Point Mamba Block}
  \begin{algorithmic}[1]
  \STATE Input: point token sequence $P_{in}: (1,N,C)$\\
  \STATE Output: point token sequence $P_{out}: (1,N,C)$\\
  \STATE \textit{/* normalize the input sequence */}
  \STATE $P'_{in}: (1,N,C) \leftarrow \text{Norm}(P_{in})$
  \STATE $x: (1,N,E) \leftarrow \text{Linear}^x(P'_{in})$
  \STATE $z: (1,N,E) \leftarrow \text{Linear}^z(P'_{in})$
  \STATE \textit{/* process with different direction */}
  \FOR{$o \in \{\text{forward}, \text{backward}\}$}
      \STATE $x_o': (1,N,E) \leftarrow \text{SiLU}(\text{Conv1d}(x))$
      \STATE $B_o: (1,N,M) \leftarrow \text{Linear}^{B_o}(x_o')$
      \STATE $C_o: (1,N,M) \leftarrow \text{Linear}^{C_o}(x_o')$
      \STATE $\Delta_o: (1,N,E) \leftarrow \log(1 + \exp(\text{Linear}^{\Delta_o}(x_o') + \text{Parameter}^{\Delta_o}))$
      \STATE $A_o: (1,N,E,M) \leftarrow A_o \otimes \text{Parameter}^{\Delta_o}$
      \STATE $B_o: (1,M,E,M) \leftarrow \Delta_o \otimes B_o$
      \STATE $y_o: (1,N,E) \leftarrow \text{SSM}(A_o, B_o, C_o)(x_o')$
  \ENDFOR
  \STATE \textit{/* get gated $y_o$ as $P_{out}$*/}
  \STATE $P^{+}_{out}: (1,N,E) \leftarrow y_\text{forward} \otimes \text{SiLU}(z)$
  \STATE $P^{-}_{out}: (1,N,E) \leftarrow y_\text{backward} \otimes \text{SiLU}(z)$
  \STATE \textit{/* residual connection */}
  \STATE $P_{hidden\_state}: (1,N,E) \leftarrow P^{+}_{out} + P^{-}_{out}$
  \STATE $P_{out}: (1,M,C) \leftarrow \text{Linear}^P(P_{hidden\_state}) + P_{in}$
  \RETURN $P_{out}$
  \end{algorithmic}
  \label{alg:1}
  \end{algorithm}

\subsection{Computation Efficiency Analysis}
\label{sec:eff}

The SSM module (shown in Fig. \ref{fig:0}) is a key part in the process of capturing global features in the entire Point Mamba, just as the core component self-attention in transformer.
Given a input point cloud $P \in \mathbb{R}^{1\times N\times C}$, the complexity of a global self-attention operation, local self-attention operation, and SSM are:
\begin{equation}
  \begin{aligned}
    &\Omega(\text{self-attention}_{\text{global}}) = O(N^2C),\\
    &\Omega(\text{self-attention}_{\text{local}}) = O(NKC),\\
    &\Omega(\text{SSM}) = O(NMC),
  \end{aligned}
  \label{eq:8}
\end{equation}
where N is the sequence length, C is the feature dimension, and K is the window size of the local self-attention operation, and M is the fixed parameter in Section \ref{sec:point mamba}.
This makes Mamba combines both the global modeling ability like global self-attention, with the linear complexity like local self-attention.

\section{Experiments}

\subsection{Dataset}
\noindent\textbf{ModelNet40 Dataset for Classification.}\hspace*{5pt} ModelNet40 \cite{4} is a widely used benchmark for 3D object classification. It contains 12,311 CAD models from 40 categories. To make a fair comparison with other classic point cloud processing backbones, we conduct experiments according to the split of 9843 models in the training set and 2468 models in the test set. We follow the same data pre-processing and data augmentation methods as PointNet \cite{10} and PCT \cite{22}, which include random anisotropic scaling for range $[\frac{2}{3}, \frac{3}{2}]$, random rotation for range $[-\pi, \pi]$, and random translation for range $[-0.2, 0.2]$.

\noindent\textbf{ScanNet Dataset for Semantic Segmentation.}\hspace*{5pt}ScanNet \cite{70} is a large-scale dataset that contains 3D scans of indoor scenes. It contains 1,513 scans from 707 indoor scenes, which are annotated with 20 parts labels. The average point number of scans in ScanNet is 148$k$. The segmentation results on the ScanNet dataset can better illustrate the advantages of Point Mamba in processing large-scale point cloud data. We follow the standard data splits in \cite{70} for training and evaluation, using 1201 scans for training and 312 scans for validation.
The data augmentation methods are consistent with OctFormer \cite{66}, which include random anisotropic scaling for range $[\frac{2}{3}, \frac{3}{2}]$, random rotation for range $[-\pi, \pi]$, random translation for range $[-0.1, 0.1]$, random elastic deformations following \cite{16} and random spherical cropping following \cite{32}.

\begin{table}[t]
  \centering
  \caption{Classification accuracy on ModelNet40. P: Points, N: Normals. All results are cited from the original papers. Point Mamba (O) and OctFormer \cite{66} share the same overall architecture. Point Mamba (C) and PCT \cite{22} share the same architecture.}
  \label{tab:1}
  \begin{tabular}{L{3.5cm}|C{1.5cm}|C{1.8cm}|C{2cm}}
  \hline
  \textbf{Method} & \textbf{Input} & \textbf{points} & \textbf{Accuracy} \\ \hline
  \end{tabular}
  \begin{tabular}{L{3.5cm}|C{1.5cm}|C{1.8cm}|C{2cm}}
  \textbf{CNN} &  &  &  \\ \hline  
  PointNet\cite{10} & P & 1k & 89.2\% \\ 
  A-SCN\cite{61} & P & 1k & 89.9\% \\ 
  SO-Net\cite{62} & P, N & 2k & 90.9\% \\ 
  PointNet++\cite{11} & P & 1k & 90.7\% \\ 
  PointNet++\cite{11} & P & 5k & 91.9\% \\ 
  PointGrid\cite{63} & P & 1k & 92.0\% \\ 
  PCNN\cite{52} & P & 1k & 92.3\% \\ 
  PointCNN\cite{5} & P & 1k & 92.5\% \\ 
  PointConv\cite{14} & P, N & 1k & 92.5\% \\ 
  P2Sequence\cite{64} & P & 1k & 92.6\% \\ 
  DGCNN\cite{13} & P & 1k & 92.9\% \\ 
  RS-CNN\cite{65} & P & 1k & 92.9\% \\ 
  PointASNL\cite{23} & P & 1k & 92.9\% \\ \hline
  \textbf{Transformer} &  &  &  \\ \hline
  NPCT\cite{22} & P & 1k & 91.0\% \\
  PCT\cite{22} & P & 1k & 93.2\% \\
  OctFormer\cite{66}&P&1k&92.7\%\\\hline
  \textbf{Mamba} &  &  &  \\ \hline
  Point Mamba (O) & P & 1k & 92.7\% \\
  Point Mamba (C) & P & 1k & \textbf{93.4\% }\\ \hline
  \end{tabular}
  \end{table}

\subsection{Implementation Details}

\noindent\textbf{Classification on ModelNet40.}\hspace*{5pt}
For classification task, we refer to the experimental settings of PCT \cite{22} and OctFormer \cite{66} and design two architectures of Point Mamba, Point Mamba (C) and Point Mamba (O), which are based on the PCT and OctFormer as the baseline respectively. The batch size in classification is 32. For Point Mamba (C), The initial learning rate is 0.01. The model is trained for 250 epochs with a cosine annealing schedule. For Point Mamba (O), The initial learning rate is 0.001. The model is trained for 200 epochs with MultiStepLR, which decays with a factor of 0.1 in the 120 and 160 epochs. All of our experiments are trained on four NVIDIA RTX 4090 GPUs.

\noindent\textbf{Semantic Segmentation on ScanNet.}\hspace*{5pt} For the segmentation task, we refer to the experimental settings of OctFormer \cite{66} and design the Point Mamba (S) architecture, which is based on the OctFormer as the baseline. We use the mean Intersection over Union (mIoU) as the evaluation metric, which is the same as OctFormer \cite{66}. Following OctFormer \cite{66}, we employ an Adam optimizer with a batch size of 16 and a weight decay of 0.05. The initial learning rate is 0.006, which decreases by a factor of 10 after 360 and epochs using the MultiStepLR scheduler. To better observe the convergence of Point Mamba, we set a training period of 800 epochs. For the input point cloud, we first apply a normalization with a scale factor of 0.01m and encoded all the points into octrees with a depth of 6. All of our experiments are trained on four NVIDIA RTX 4090 GPUs.

\subsection{Classification Results on ModelNet40 dataset} Tab. \ref{tab:1} shows the classification accuracy of Point Mamba and other point cloud processing backbone on ModelNet40. Our experimental results show that the classification accuracy of Point Mamba (C) on the ModelNet40 dataset is 93.4\%, which has surpassed our baseline PCT \cite{22} of 93.2\%. Point Mamba (O) achieves 92.7\% accuracy, which is also competitive with the baseline OctFormer \cite{66} of 92.7\%.

\begin{table}[t]
  \centering
  \caption{Semantic segmentation accuracy on ScanNet dataset. Val. denotes the mIoU on the validation set. All results are cited from the original papers.}
  \label{tab:2}
  \begin{tabular}{L{5cm}|C{3cm}}
  \hline
  \textbf{Method} & {Val.}\\ \hline
  \end{tabular}
  \begin{tabular}{L{5cm}C{3cm}}
  \textbf{CNN} &  \\ \hline
  \end{tabular}
  \begin{tabular}{L{5cm}|C{3cm}}
  PointNet++\cite{11} & 53.5\% \\
  JointPoint\cite{74} & 69.2\% \\
  PointConv\cite{14} & 61.0\% \\
  PointASNL\cite{23} & 63.5\% \\
  SparseConvNet\cite{17} & 69.3\% \\
  MinkowskiNet\cite{16} & 70.1\% \\
  O-CNN\cite{75} & 74.4\% \\\hline
  \end{tabular}
  \begin{tabular}{L{5cm}C{3cm}}
  \textbf{Transformer} &  \\ \hline
  \end{tabular}
  \begin{tabular}{L{5cm}|C{3cm}}
  Point Transformer\cite{22} & 70.6\% \\
  Fast Point Transformer\cite{28} & 72.1\% \\
  Stratified Transformer\cite{32} & 74.3\% \\
  Point Transformer V2\cite{22} & 75.4\% \\
  OctFormer\cite{66} & 74.5\%  \\ 
  OctFormer (voting)\cite{66} &\bf{75.7\%} \\ \hline
  \end{tabular} 
  \begin{tabular}{L{5cm}C{3cm}}
  \textbf{Mamba} &  \\ \hline
  \end{tabular}
  \begin{tabular}{L{5cm}|C{3cm}}
  Point Mamba & \textbf{74.6\%} \\
  Point Mamba (voting)&\bf{75.7\%}\\ \hline
  \end{tabular}
  \end{table}
  
\subsection{Segmentation Results on ScanNet Dataset}
\begin{figure}[t]
  \centering
  \includegraphics[width=0.7\textwidth]{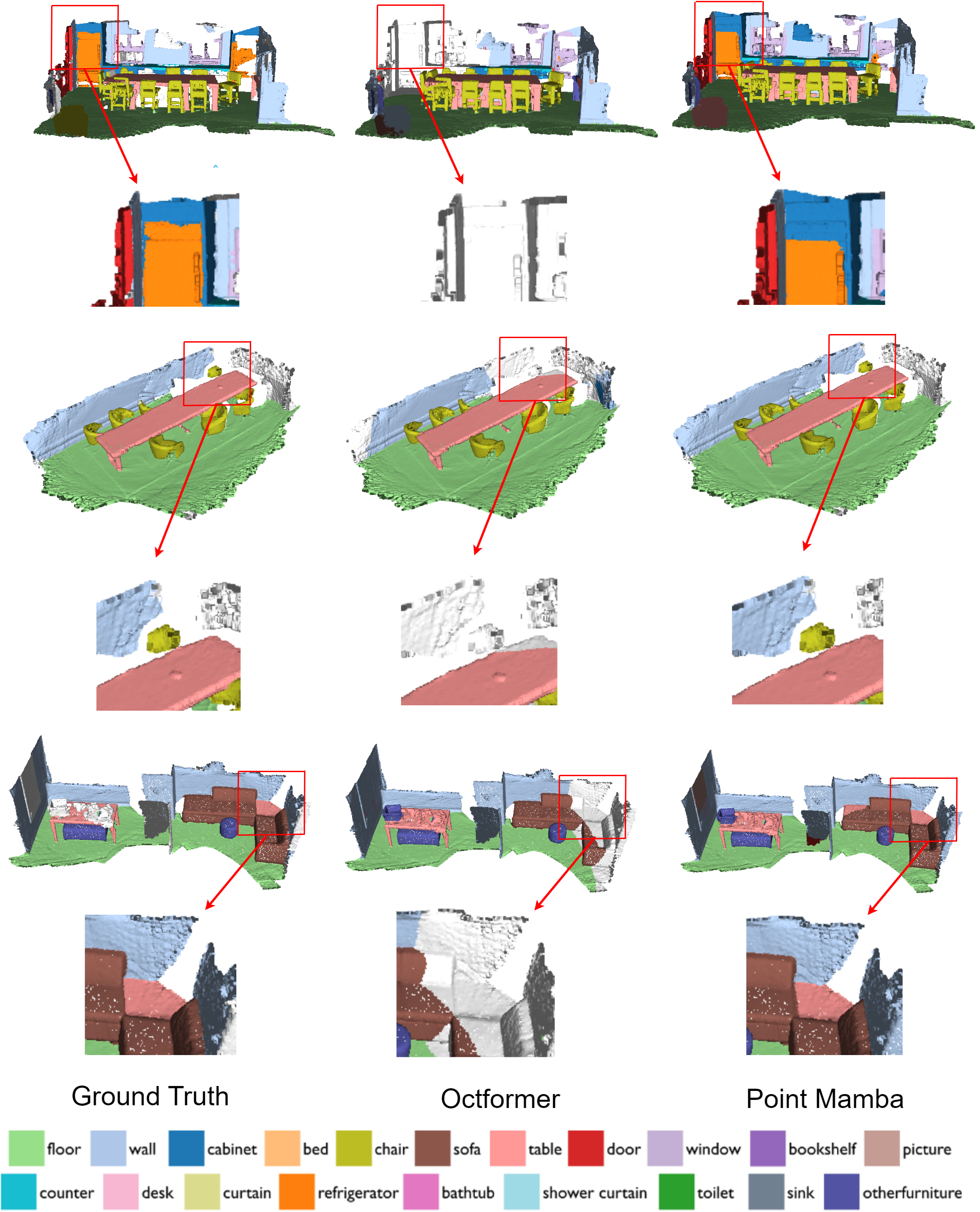}
  \caption{The segmentation results on the ScanNet dataset. The first column is the ground truth, the second column is the prediction of OctFormer, and the third column is the prediction of Point Mamba. The global modeling ability of our Point Mamba improves the recognition and segmentation of consistent semantic objects, like the walls.}
  \label{fig:2}
\end{figure}

Our experimental results in Tab. \ref{tab:2} shows that before voting our Point Mamba achieves 74.6 mIOU in ScanNet, which is higher than the OctFormer \cite{66} of 74.5 mIOU. Considering that Stratified Transformer \cite{32}, Point Transformer V2 \cite{22}, and OctFormer \cite{66} use a strategy of voting to improve the performance, we also use the same strategy to improve the performance of Point Mamba. After voting, the mIoU of Point Mamba is 75.7\%, which achieves a competitive performance with our baseline on the ScanNet dataset. Fig. \ref{fig:2} shows several semantic segmentation examples on the ScanNet dataset. The first column is the ground truth label. The second column is the prediction for OctFormer. The third column is the prediction of our Point Mamba. It is obvious that our Point Mamba has better segmentation performance since the global modeling on the whole input points can enable the consistent semantic region recognition.

\subsection{GPU and Parameter Efficiency}
\noindent\textbf{Parameters and FLOPs.}\hspace*{5pt} Tab. \ref{tab:3} shows the comparison of learnable parameters and FLOPs between Point Mamba and OctFormer, which both have linear complexity. All parameters are tested on an 80GB NVIDIA A100. For the number of parameters, Point Mamba has over 25\% fewer parameters than OctFormer in both classification and semantic segmentation tasks. For FLOPs, we notice that Point Mamba has slightly more FLOPs than OctFormer for the simpler classification task, but for the semantic segmentation task, Point Mamba has fewer FLOPs than OctFormer.
\begin{table}[tb]
  \centering
  \caption{Comparison of parameters and FLOPs between Point Mamba and OctFormer.}
  \label{tab:3}
  \begin{tabular}{L{5cm}|C{3cm}|C{3cm}}
  \hline
  \textbf{Method} & \textbf{Parameters} & \textbf{FLOPs} \\ \hline
  \end{tabular}
  \begin{tabular}{L{5cm}|C{3cm}|C{3cm}}
  OctFormer (cls) &4.04M  & 1.28G \\ \hline
  Point Mamba (cls) &3.08M  &1.31G \\\hline
  \end{tabular}
    \begin{tabular}{L{5cm}|C{3cm}|C{3cm}}
    OctFormer (seg) &  44.03M&56.52G   \\ \hline
    Point Mamba (seg) & 31.99M& 55.07G \\ \hline
  \end{tabular}
  \end{table}
  
\begin{figure}[tb]
  \centering
  \includegraphics[width=0.8\textwidth]{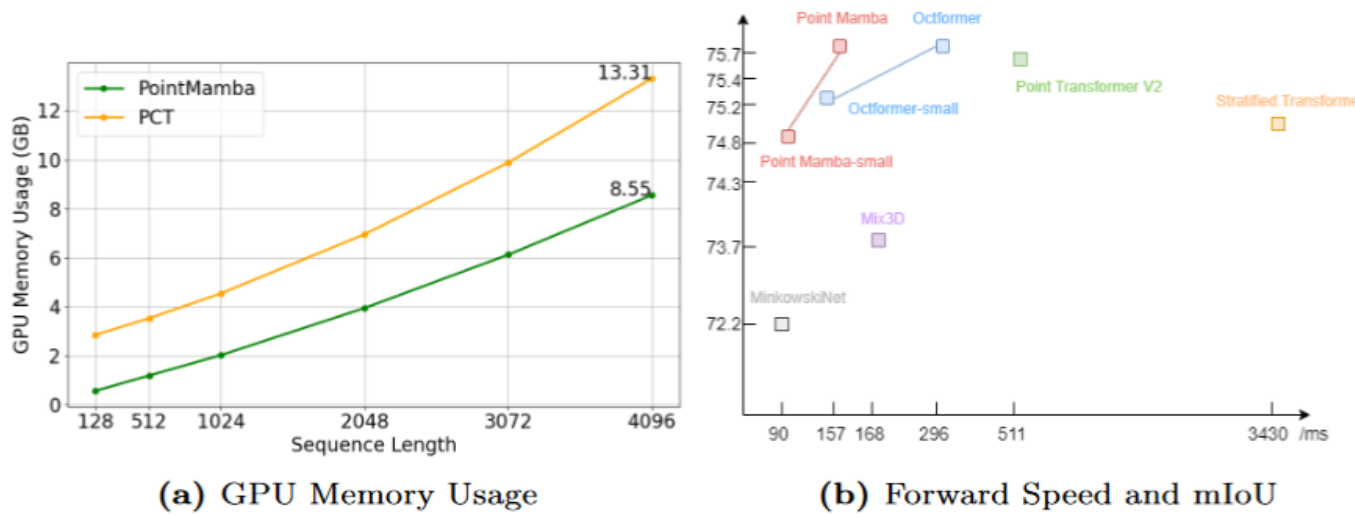}
  \caption{Fig. (a) shows the GPU memory usage of Point Mamba and PCT with a varying sequence length. 
  Point Mamba retains linear memory usage and has lower memory usage than PCT. Fig. (b) shows the forward speed and mIoU of Point Mamba, voxel-based CNNs, and transformer backbones on ScanNet. Point Mamba achieves a competitive mIoU with a faster forward speed.}
  \label{fig:3}
\end{figure}

\noindent\textbf{GPU Efficiency.}\hspace*{5pt} As shown in Fig. \ref{fig:3}, we compare the GPU memory usage of Point Mamba and PCT \cite{22}, which are evaluated on an NVIDIA A100. The results show that with the growth of the sequence, the memory usage of Point Mamba remains linear and has lower memory usage than PCT.

\noindent\textbf{Inference Time.}\hspace*{5pt} Fig. \ref{fig:3} shows the forward speed and mIoU on ScanNet of Point Mamba and other voxel-based CNNs and transformer backbones. Point Mamba-small achieves a competitive mIoU of 74.8\% after voting with the fastest forward speed of 90ms. This significantly demonstrates the performance advantage of Point Mamba over its counterparts.

\subsection{Ablation Study}

\subsubsection{Shuffle Key Order:}
In section \ref{octree}, we introduce the shuffle key order to construct the octree, which determines the z-order causality of the ordering points.
In our default setting, the key order is set to ($x,y,z$). In this experiment, we will compare five other key orders, including ($y,x,z$), ($z,x,y$), ($z,y,x$), ($y,z,x$), and ($x,z,y$).
The results are shown in Tab. \ref{tab:4left}.

\begin{table}[tb]
  \centering
  \begin{minipage}[t]{0.45\textwidth}
    \centering
    \caption{Ablation study of shuffle key order.}
    \label{tab:4left}
    \begin{tabular}{C{3cm}|C{2cm}}
      \hline
      \textbf{Shuffle Key Order} & \textbf{Accuracy} \\ \hline
      $x,y,z$ & \bf{92.7\%} \\
      $y,x,z$ & 92.2\% \\
      $z,x,y$ & 92.3\% \\
      $z,y,x$ & 92.1\% \\
      $y,z,x$ & \bf{92.7\%} \\
      $x,z,y$ & 92.2\% \\ \hline
    \end{tabular}
  \end{minipage}
  \hfill
  \begin{minipage}[t]{0.45\textwidth}
    \centering
       \caption{Ablation study of different Octree depths.}
    \label{tab:5left}
    \begin{tabular}{C{3cm}|C{2cm}}
      \hline
      \textbf{Octree Depth} & \textbf{Accuracy} \\ \hline
      5 & 91.9\% \\
      6 & \bf{92.7\%} \\
      7 & 92.0\% \\ \hline
    \end{tabular}

  \end{minipage}
\end{table}

\subsubsection{Octree Depth}
The construction of the octree is aimed at building the causal dependence of the point cloud and forming a memory mapping that is conducive to access. The depth of the tree is an important parameter. We compare the performance of octrees with different depths in Tab. \ref{tab:5left}.

\begin{table}[tb]
  \centering
  \begin{minipage}[t]{0.45\textwidth}
    \centering
            \caption{Ablation study of different classification block channels.}
    \label{tab:5right}
    \begin{tabular}{C{3cm}|C{2cm}}
      \hline
      \textbf{Block Channel} & \textbf{Accuracy} \\ \hline
      (96,96) & 92.1\% \\
      (96,192) & \bf{92.7\%} \\
      (96,384) & 92.3\% \\ \hline
    \end{tabular}
 
  \end{minipage}
  \hfill
  \begin{minipage}[t]{0.45\textwidth}
    \centering

            \caption{Ablation study of different segmentation block channels.}
    \label{tab:4right}
    \begin{tabular}{C{3cm}|C{2cm}}
      \hline
      \textbf{Model Scale} & \textbf{Accuracy} \\ \hline
     (2,2,6,2) & 73.3\% \\
     (2,2,6,2) & 74.8\%(voting) \\ 
      (2,2,18,2) & \bf{74.6\%} \\
      (2,2,18,2)& \bf{75.7\%}(voting)\\
      \hline
    \end{tabular}

  \end{minipage}
\end{table}

\subsubsection{Number of Channels:}
The number of channels in each Block of Point Mamba and Point Mamba Blocks is an important indicator of the model size. We compare the performance of Point Mamba with different numbers of channels, including classification and segmentation tasks. For the classification task, we use ($C_1$,$C_2$) to represent the number of channels in two block sets with 6 blocks each. For the segmentation task, we use ($N_1$,$N_2$,$N_3$,$N_4$) to represent the number of blocks in four stages with Channels set (96, 192, 384, 384). Results are shown in Tab. \ref{tab:5right} and Tab. \ref{tab:4right}.

\section{Limitation and Future Work}
The point order sequence in our work is shuffled by octree keys. In the future, we will delve deeper into whether exists more effective point sorting methods. Point mamba with larger sizes and parameters can possess more powerful ability. We leave this for the future exploration. Furthermore, more 3D deterministic large-scale tasks using point mamba architecture are underdeveloped. Due to the linear complexity, the extension to large-scale with high efficiency is promising.

\section{Conclusion}
In this paper, we propose a novel point cloud processing backbone, Point Mamba, which is based on the state space model (SSM). By introducing the octree-based ordering scheme, we re-organize the disorder of the point cloud and form a z-order causal sequence that is adapted to the Mamba backbone. Then, we design the Point Mamba block, which combines the bidirectional selective scanning mechanism to adjust the sequence-order dependence of Mamba. Our experimental results on the ModelNet40 and ScanNet datasets show that Point Mamba has comparable performance to the transformer-based backbone in point cloud processing tasks.

\bibliographystyle{unsrt}  
\bibliography{references}

\end{document}